\documentclass[sigplan,screen,authorversion,nonacm]{acmart}

\AtBeginDocument{%
  \providecommand\BibTeX{{%
    \normalfont B\kern-0.5em{\scshape i\kern-0.25em b}\kern-0.8em\TeX}}}

\usepackage{color}
\usepackage{csvsimple}

\newcommand{\vansh}[1]{{\color{magenta}~{\em Comment by Vansh: #1}}}

\usepackage{subcaption}
\usepackage{multirow}
\begin{document}

\title{BEACON: Balancing Convenience and Nutrition in  Meals With Long-Term Group Recommendations and Reasoning on Multimodal Recipes}

\author{Vansh Nagpal}
\email{vnagpal@email.sc.edu}
\affiliation{%
  \institution{University of South Carolina}
  \city{Columbia}
  \state{SC}
  \country{USA}
}

\author{Siva Likitha Valluru}
\email{svalluru@email.sc.edu}
\affiliation{%
  \institution{University of South Carolina}
  \city{Columbia}
  \state{SC}
  \country{USA}
}

\author{Kausik Lakkaraju}
\email{kausik@email.sc.edu}
\affiliation{%
  \institution{University of South Carolina}
  \city{Columbia}
  \state{SC}
  \country{USA}
}

\author{Biplav Srivastava}
\email{biplav.s@sc.edu}
\affiliation{%
  \institution{University of South Carolina}
  \city{Columbia}
  \state{SC}
  \country{USA}
}




\begin{abstract}
A common, yet  regular, decision made by people, whether healthy or with any health condition, is to decide what to have in meals like breakfast, lunch, and dinner, consisting of a combination of foods for appetizer, main course, side dishes, desserts, and beverages. However, often this decision is seen as a trade-off between nutritious choices (e.g., low salt and sugar) or convenience (e.g., inexpensive, fast to prepare/obtain, taste better). In this preliminary work, we present a data-driven approach for the novel meal recommendation problem that can explore and balance choices for both considerations while also reasoning about a food's constituents and cooking process. Beyond the problem formulation, our contributions also include a goodness measure, a recipe conversion method from text to the recently introduced multimodal rich recipe representation (R3) format, and learning methods using contextual bandits that show promising results. 

\end{abstract}








\maketitle
\section{Introduction}

Although it is well known that  nutritious foods are essential to a person's health, the actual adherence to dietary requirements is quite poor across the world. In fact, according to a recent meta-survey \cite{leme2021adherence}, almost 40\% of population across high and low-and-medium income countries do not adhere to their national food-based dietary guidelines, often prioritizing convenience over nutrition needs.
A complete meal typically consists of an appetizer, main course, side dish, dessert, and a beverage. Some people prefer getting food recommendations from their friends or family, and others turn to online recommender systems \cite{yang2017yum} or even Large Language Models (LLMs) as they have become easily available in the form of chatbots. But even when users adhere to recommendation, the systems can be inaccurate.  Authors in \cite{papastratis2024can} found assessing ChatGPT-based recommenders for balanced diets in non-communicable diseases (NCDs)  patients. 

\begin{figure}[h]
    \centering
    \includegraphics[width=0.99\linewidth]{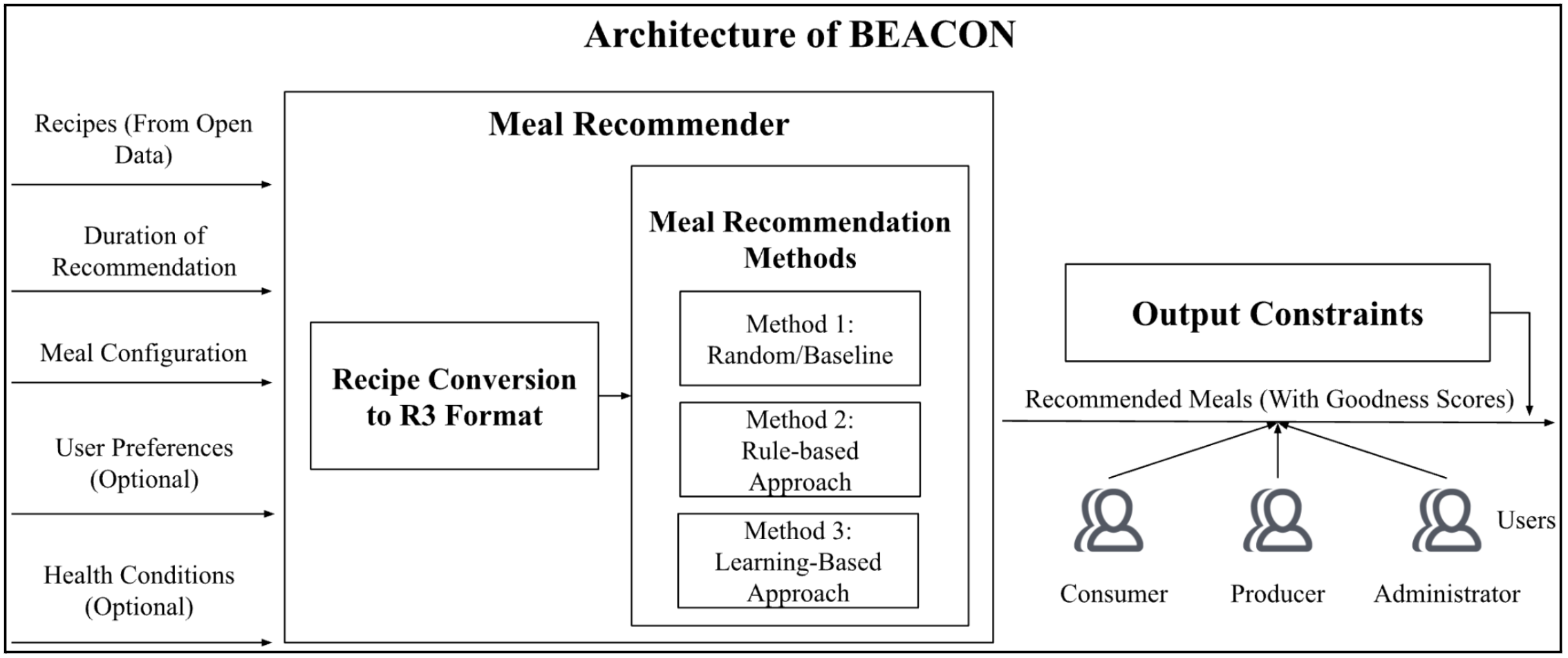}
    \caption{A brief depiction of BEACON's architecture.}
    \label{fig:archi}
\end{figure}

We seek to help the general population decide meal choices while nudging them towards healthy choices by leveraging data from online recipes, domain knowledge about meals and how they are configured from foods, and user preferences (Figure~\ref{fig:archi}).
In doing so, we recognize the reality that people want to explore a variety of foods, and a {\em long-horizon meal recommender} can act as a trusted companion seeking to keep the user well informed even when they deviate from nutrition guidelines. 



Our contributions are that we: (1) introduce the novel meal recommendation problem needing food choices over meal configurations and long time horizons, (2) present the  BEACON\footnote{BEACON stands for \underline{B*A}lancing \underline{CO}nvenience and \underline{N}utrition Meals With Long-Term Group Recommendations and Reasoning on Multimodal Recipes} system to address the problem of meal recommendation over long periods and provide a case study to motivate its usage, (3) adopt the multimodal R3 format and convert fast food recipes from two well-known fast food chains: Taco Bell and McDonald's into R3 representations, using LLMs augmented with human supervision, (4) introduce novel quantitative and qualitative metrics to evaluate the recommendation to measure the duplicates, coverage, and user constraint satisfaction, and (5) show the effectiveness of the meal recommendations over suitable baselines.

In the remainder of the paper, we begin with a background on automated recommendations of healthy and personalized meal plans, and recipe representations, followed by the problem formulation, system implementation, and recommendation and recipe evaluation results. We conclude with a summary of future work.

\vspace{-1em}
\section{Background and Related Work}

There is a large literature on recommendation methods for single items \cite{su2009survey,cremonesi2010performance}. In many practical situations,  a \textit{group} of items have to be recommended, like forming teams. Here, although the group problem can be treated as   a special case of \textit{sequential} single-item recommendation problem \cite{srivastava2022ultra}, better results are found when treating them as a group \cite{valluru2024promoting, valluru2024ultra}. BEACON falls in the latter category.


There are food recommendation systems in literature which seek to 
guide users based on their dietary preferences, health conditions, and nutritional requirements. 
They offer tailored guidance on nutritionally balanced meal options, ensuring that individuals consume the right combination of macronutrients and micronutrients to support their physical activity and overall health~\cite{bekdash2024epigenetics}. 
Examples are \textit{single} food items~\cite{yang2017yum, ge2015using}.
They may also help with weight management, such as portion control and unhealthy food cravings~\cite{dunn2018mindfulness}, by offering satisfying yet calorie-conscious alternatives. 

Closest to our work is the SousChef system~\cite{ribeiro2017souschef}, which tackles 
the problem of food recommendation\footnote{They call it meals but do not have the rich meal configurations we support.} for improving the health of older adults. SousChef utilizes a two stage rule-based algorithm to filter away incompatible choices and then recommends food  using data about user preferences and item ingredients to recommend multi-item food plans based on nutritional needs; we consider a longer horizon problem setting, learning based methods and a multimodal dataset. Additionally, other systems~\cite{zeevi2015personalized, forouzandeh2024health, min2019food, zioutos2023healthy, pecune2020recommender} have explored various aspects of food recommendation but often do not incorporate a structured representation that includes both the food content and preparation process.


\section{Problem}

BEACON processes several inputs, including the R3 Food Recipe Set ($\mathcal{R}$), and user information ($\mathcal{U}$) which includes dietary conditions categorized as either Healthy or Diabetic. Additionally, recommendation constraints ($\mathcal{C}$) are defined to tailor the meal plans. These constraints include DailyMeal categories (Breakfast, Lunch, Dinner), MealType configurations (MealConfig and Time), and MealConfig components (Beverage, Main, Side, Dessert). 
Specifically, we configure all recommendations to be in the following format (per day): Breakfast(Main Course, Beverage), Lunch(Main Course, Side, Beverage), and Dinner(Main Course, Side, Dessert, Beverage).  The time period for recommendations is set to a configurable duration length; for simulation reasons, we consider a minimum of 1 day and maximum length of 5 days. BEACON also introduces novel quantitative and qualitative metrics to evaluate its recommendations, measuring aspects such as duplicates, coverage, and user constraint satisfaction. 
Our approach ensures that the recommendations are tailored to individual needs, promoting better health management and dietary adherence. 

\subsection{Use Cases}
One key use case of BEACON involves diabetic individuals from minority communities who need culturally relevant and nutritionally balanced meal plans to manage their blood sugar levels. For instance, BEACON can recommend a complete meal that aligns with traditional dietary practices while ensuring low glycemic index foods are included. Another use case involves busy professionals seeking convenient, healthy meal options that fit into their hectic schedules~\cite{brown2005college, pelletier2012balancing, king2021too}. Restaurants and catering services can also use BEACON to create diverse and balanced group meal plans that meet the dietary needs of various customers. Lastly, BEACON can also be a valuable tool for public health professionals working in school or hospital settings. In schools, BEACON can help design balanced and appealing meal plans that meet nutritional standards and cater to diverse student preferences, promoting healthy eating habits from a young age~\cite{vieux2013dietary}. In hospitals, BEACON can assist dietitians in creating meal plans for patients with specific dietary requirements, ensuring that meals are not only nutritionally adequate but also appetizing and easy to prepare~\cite{iff2008meeting, barocco2019evaluation}.
\section{System Implementation}
Figure \ref{fig:archi} shows the proposed BEACON's architecture. 

\begin{table}[ht]
{\tiny
    \centering
    \begin{tabular}{|p{5em}|p{4.5em}|p{5em}|p{5em}|p{7em}|}
    \hline
     \textbf{Category} & 
     \textbf{hasNuts \%} & 
     \textbf{hasMeat \%} & 
     \textbf{hasDairy \%} &
     \textbf{Total Recipes}
     \\ \hline
    McDonald`s & 9.10 & 63.63 & 90.90 & 11 \\ \hline
    TREAT Recipes \cite{pallagani2022rich} & 17.24 & 34.48 & 48.28 & 29 \\ \hline
    Taco Bell & 0 & 60 & 100 & 10 \\ \hline
    \end{tabular}
    \caption{Table showing the \% of recipes with nuts (hasNuts), with meat (hasMeat), with dairy products (hasDairy), and the total number of recipes under each category.}
    \label{tab:recipe-stats}
}
\end{table}
\subsection{Data: Recipes and R3 Preparation}

The existing recipes on the internet are available as textual documents which makes it difficult for machines to read and reason. Better representation of such information can improve decision support systems and also provide an easy way to query and get insights from the data. \cite{pallagani2022rich} introduced a Rich Recipe Representation (R3) in the form of 
Planning Domain Definition Language (PDDL) plans. They created 25 egg-based recipes in R3 manually from original recipes taken from the RecipeQA dataset \cite{yagcioglu2018recipeqa}. By combining R3 representation, multi-modal content extraction, and group recommendation techniques, our work addresses diverse user needs, from individuals with specific dietary preferences to businesses catering to group dining experiences.

In addition, we considered fast food recipes which are known for their convenient access.
We generate  recipes for 11 items served by McDonald's and 10 served by Taco Bell. To generate data in a time-effective manner, we use an GPT-3.5 to convert the recipes into their intended R3 structure. As LLMs are known for hallucinating, we manually verify each recipe for quality assurance. For extracting ingredients with amounts and units, we use 0-shot prompting. For properly extracting instructions with atomic steps, we utilize the chain-of-thought method with few-shot prompting~\cite{ma2024fairness, zhang2023machine} by first inputting to the LLM a single instruction and its corresponding R3 breakdown. We then generate the R3 representation for an entire recipe by giving the numbered instructions from the recipe as input. We do this for 21 fast food recipes to generate 21 corresponding R3 representations. We also add 4 non-fast food recipes to the original 25 recipes from \cite{pallagani2022rich} to achieve a total of 50 R3 generated recipes. We also manually annotate each item's R3 representation with 
its binary food features (hasDairy, hasNuts, hasMeat). Table \ref{tab:recipe-stats} shows the \% of recipes with nuts, meat, dairy products, and the total number of recipes under each category.

\noindent\textbf{Goodness Metric:} We use 3 different evaluation criteria to decide the goodness of a recommendation. These criteria include assessing the recurrence of items across meals and within meals (duplicate metric, $md$), assessing if the recommended meals satisfy the preferred components (meal coverage metric, $cs$), and assessing if the recommended meals have ingredients that match user preferences (user-constraint metric, $uc$). 
For each recommendation, we compute the goodness score $G$ as a weighted sum of the individual scores $md$, 
$cs$, and $uc$, with weights tailored to user preferences (e.g., some users may prefer having duplicates in their meals, while others may not).
We elaborate on the three metrics below:

\noindent(1) \textbf{Duplicate Metric. }
Our duplicate metric examines the occurrence of repeated items within meal recommendations in two distinct ways. Firstly, we consider a repeated item within a meal as a \textit{meal item duplicate,} and secondly, we check the repetition of items across different meals in the same day, termed as a \textit{role item duplicate.} The meal item duplicate score, denoted as $md_i$ for a particular meal $m_i$, measures the ratio of unique items to total items within the meal. For a recommendation comprising multiple meals $m_1, ..., m_n$, we calculate the meal duplicate score $md$ as the average of all meal item duplicate scores $md_1, ..., md_n$. 

\noindent(2) \textbf{Meal Coverage Metric.} Our meal coverage metric evaluates the extent to which a meal recommendation aligns with the user's desired food roles. To begin, we first establish a meal configuration that outlines the typical roles found in a meal: main course, side dish, dessert, and beverage. Each role is assigned an ideal weight, allowing users to prioritize certain roles over others based on their preferences. We then annotate our dataset utilizing GPT-3.5 to assign food items an array of their appropriate roles.

To calculate the coverage score for a recommendation, we analyze the presence of recommended meal items corresponding to their role and alignment with user preferences. If a recommended item matches its assigned role and aligns with user preferences (indicated using a weight of $+1$), it positively contributes to the coverage score. Conversely, misaligned  recommended items (e.g., recommending a beverage item like soda as a side dish), or items that the user has explicitly stated they do not want (e.g., avoiding nuts or dairy due to allergen requirements) would incur a penalty on the coverage score. For each meal, $m_i$, we therefore calculate a coverage score $cs_i$ by taking the ratio of requested roles fulfilled to the number of requested roles. We calculate our final coverage score, $cs$, as the average of all scores $cs_1, ..., cs_n$.

\noindent(3) \textbf{User Constraint Metric.} In addition to specifying the types of food roles they prefer in each meal, users also provide their ingredient preferences, focusing on three key features: dairy content, meat content, and nuts content. These features were chosen to create a minimally functional system, with plans to extend the list of features in the future. Each feature can have a user preference value of $-1$, $0$, or $+1$, representing a negative preference (the user prefers meals without this ingredient), a neutral preference (the user is indifferent), and a positive preference (the user prefers meals with this ingredient), respectively. Our system is designed to be flexible, allowing any number of such features to be added or removed based on user requirements. 

We manually annotate each of our $50$ R3 representations with corresponding feature flags, indicating whether an item contains a particular ingredient (e.g., dairy, meat, or nuts). For each meal $m_i$, we calculate a user constraint score $uc_i$ by comparing the user's preference with the meal's ingredient content. If the user's preference is negative ($-1$) and the meal contains the ingredient, this counts negatively towards the score. Conversely, if the user's preference is positive ($+1$) and the meal contains the ingredient, this counts positively towards the score. Neutral preferences ($0$) do not affect the score regardless of the meal's content. Additionally, if a meal does not contain an ingredient that the user positively prefers, the overall goodness is not penalized unless the user specifies it through a configurable flag.


We calculate the final user constraint score \( uc \) as the average of $uc_1, uc_2, ..., uc_n$ across all meals in the recommendation.

\vspace{-0.7em}
\subsection{Meal Recommendation}
We used three different methods to recommend meals to users: (1) \textit{M0: Random/Baseline}, (2) \textit{M1: Sequential}, and (3) \textit{M2: Relational Boosted Bandit}.

\textit{M0} serves as our baseline method for meal recommendations. In this approach, meals are generated randomly i.e., the selection of items for each meal does not consider any user preferences, dietary restrictions, or allergen information. An advantage of \textit{M0} is that it has the potential to introduce users to a diverse array of food choices. By forming meals without any specific criteria, users might encounter new and varied food items that they may not have otherwise considered, thereby broadening their culinary experiences.

\textit{M1} introduces a more structured approach compared to \textit{M0}. In this method, we use our dataset of recipes and rotate through them to recommend meals. This sequential nature ensures that each recipe in the list is eventually recommended, however, like \textit{M0}, \textit{M1} does not take into account any individual user needs. Its primary advantage over \textit{M0} is only the avoidance of repetitive randomness.

\textit{M2} represents a significant advancement over both \textit{M0} and \textit{M1} by incorporating user preferences, dietary restrictions, allergen information, and item roles (e.g., dessert, main course) into the meal recommendation process. Unlike the previous methods, \textit{M2} uses contextual bandits and reinforcement learning to automatically extract and learn complex rules from data~\cite{kakadiya2021relational}, allowing it to provide highly personalized meal recommendations. \textit{M2} also continuously learns and improves its recommendations over time as more data is provided, making it a dynamic and evolving system. This method was adapted from an effective approach used in a recent work~\cite{valluru2024promoting, valluru2024ultra} from a different domain, where the goal was to recommend teams that could successfully complete an RFP (Request for Proposals).

\section{Recommendation Evaluation}
We conduct our experiments for each recommendation method across 3 different time frames $t_1$~(1 day) $t_2$~(3 days), and $t_3$~(5 days) and 3 user configurations, $c_1, c_2,$ and $c_3$. These configurations all include 24 users. For each of the food features that we consider (hasDairy, hasMeat, hasNuts), we consider corresponding user features $\{u_f\}$
that can take on a value of -1, 0, or 1. These values correspond to negative preference, neutral preference, and positive preference respectively. For each $c_i$ and feature $u_f$, we randomly select $p_i$ users to have a positive preference to $u_f$, $n_i$ users to have a negative preference to $u_f$, and the remaining to have a neutral preference to $u_f$. In $c_1$, $c_2$, and $c_3$, we choose $n_1 = p_1 = 12$, $n_2 = p_2 = 8$, and $n_3 = p_3 = 2$. Thus, the constraints  on users preferences are decreasing across the configurations.
Each $c_i$ is referred to as $n_i$/(24 - $n_i$ - $p_i$)/$p_i$ in Table ~\ref{tab:results_table}, corresponding to negative, neutral, and positive preference. For each experiment that we conducted, we display our 3 metrics: user constraint ($uc$), duplicate meal ($dm$), and meal coverage ($mc$), as well as their various combinations calculated as averages. We display our results in Table ~\ref{tab:results_table}.

As shown in Table ~\ref{tab:results_table}, the bandit algorithm outperforms other methods in the user constraint metric and meal coverage metric, which is expected because it is the most informed out of the three. It is important to note that when there are fewer users with negative preferences towards the food features, the random and sequential algorithms only perform marginally worse because users are less particular about their preferences. The sequential algorithm scores perfectly in the duplicate meal metric because there are more items in our dataset than are in a meal, while the bandit algorithm performs the worst, which is caused by the bandit favoring very few items with a higher probability of being a positively recommended item. This causes the bandit algorithm to perform poorly in the $uc\_dm$ combination metric for most trials. Additionally, in the $uc\_dm$ metric, the bandit performs the worst in the $c_3$ configuration because users are less particular and the bandit is more likely to recommend duplicates. However, since, the random and sequential algorithms don't perform nearly as well in the $mc$ and $uc$ metrics, the bandit algorithm performs significantly better in the $uc\_dm\_mc$, $uc\_mc$, and $dm\_mc$ combination metrics.\\

\begin{table}[ht]
\centering
{\tiny
\begin{tabular}{|p{3em}|p{4em}|p{1.7em}|c|p{1.8em}|c|c|p{2.2em}|p{2.4em}|}
\hline
\textbf{Config} & 
\textbf{Algorithm} & 
\textbf{uc} & 
\textbf{dm} & 
\textbf{mc} & 
\textbf{uc\_dm\_mc} & 
\textbf{uc\_dm} & 
\textbf{uc\_mc} & 
\textbf{dm\_mc} 
\\ \hline

\multirow{3}{3em}{$c_1, t_1$}    
& bandit 
& \textbf{0.875}
& 0.890 
& \textbf{0.993} 
& \textbf{0.919}
& 0.883 
& \textbf{0.934} 
& \textbf{0.942} 
\\ \cline{2-9}

& 
sequential & 
0.806 & 
\textbf{1.000} & 
0.384&  
0.730 & 
\textbf{0.903}& 
0.595& 
0.692 
\\ \cline{2-9}

& 
random&
0.736& 
0.978& 
0.454& 
0.723& 
0.857& 
0.595& 
0.716 
\\ \hline
 
\multirow{3}{3em}{$c_1, t_2$} & 
bandit & 
\textbf{0.870}& 
0.905 & 
\textbf{0.984} & 
\textbf{0.920}& 
0.888 & 
\textbf{0.927} & 
\textbf{0.944} 
\\ \cline{2-9}

& 
sequential & 
0.806 & 
\textbf{1.000} & 
0.380 & 
0.729 & 
\textbf{0.903} & 
0.593 & 
0.690 
\\ \cline{2-9}

& 
random & 
0.779 & 
0.995 & 
0.438 & 
0.737 & 
0.887 & 
0.608 & 
0.716 
\\ \hline

\multirow{3}{3em}{$c_1, t_3$} & 
bandit & 
\textbf{0.918} &
0.914 & 
\textbf{0.993} & 
\textbf{0.942} & 
\textbf{0.916} & 
\textbf{0.955} & 
\textbf{0.954} 
\\ \cline{2-9}

& 
sequential & 
0.796 & 
\textbf{1.000} & 
0.377 & 
0.725 & 
0.898 & 
0.587 & 
0.689 
\\ \cline{2-9}

& 
random & 
0.775 & 
0.995 & 
0.439 & 
0.736 & 
0.885 & 
0.607 & 
0.717 
\\ \hline

\multirow{3}{3em}{$c_2, t_1$} & 
bandit & 
\textbf{0.954}& 
0.918 & 
\textbf{0.986} & 
\textbf{0.953} & 
\textbf{0.936} & 
\textbf{0.970} & 
\textbf{0.952} 
\\ \cline{2-9}

& 
sequential & 
0.852 & 
\textbf{1.000} & 
0.384& 
0.745& 
0.926& 
0.618& 
0.692 
\\ \cline{2-9}

& 
random& 
0.847& 
0.988& 
0.400& 
0.745& 
0.918& 
0.624& 
0.694 
\\ \hline

\multirow{3}{3em}{$c_2, t_2$} & 
bandit & 
\textbf{0.948}& 
0.949 & 
\textbf{0.977} & 
\textbf{0.958} & 
\textbf{0.948} & 
\textbf{0.962} & 
\textbf{0.963} 
\\ \cline{2-9}

& 
sequential& 
0.861 & 
\textbf{1.000}& 
0.380& 
0.747& 
0.931& 
0.621& 
0.69 
\\ \cline{2-9}

& 
random& 
0.856& 
0.996& 
0.412& 
0.755& 
0.926& 
0.634& 
0.704 
\\ \hline

\multirow{3}{3em}{$c_2, t_3$} & 
bandit & 
\textbf{0.917}& 
0.933 & 
\textbf{0.949} & 
\textbf{0.933}& 
0.925 & 
\textbf{0.933} & 
\textbf{0.941} 
\\ \cline{2-9}

& 
sequential& 
0.869 & 
\textbf{1.000}& 
0.377& 
0.749 & 
\textbf{0.934}& 
0.623& 
0.689 
\\ \cline{2-9}

& 
random & 
0.851& 
0.992& 
0.437& 
0.760 & 
0.922& 
0.644& 
0.715 
\\ \hline

\multirow{3}{3em}{$c_3, t_1$} & 
bandit & 
\textbf{0.986}& 
0.954 & 
\textbf{1.000} & 
\textbf{0.980}& 
0.970 & 
\textbf{0.993} & 
\textbf{0.977} 
\\ \cline{2-9}
& 
sequential& 
0.968 & 
\textbf{1.000}&
0.384& 
0.784 & 
\textbf{0.984}& 
0.676& 
0.692 
\\ \cline{2-9}
& 
random & 
0.963& 
0.984& 
0.447& 
0.798& 
0.973&
0.705& 
0.715 
\\ \hline

\multirow{3}{3em}{$c_3, t_2$} & 
bandit & 
\textbf{0.986}& 
0.951 & 
\textbf{0.993} & 
\textbf{0.977}& 
0.968 & 
\textbf{0.990} & 
\textbf{0.972} 
\\ \cline{2-9}

& 
sequential& 
0.961 & 
\textbf{1.000}& 
0.380 & 
0.781 & 
\textbf{0.981}& 
0.671& 
0.690 
\\ \cline{2-9}

& 
random& 
0.960& 
0.991& 
0.411& 
0.787& 
0.975& 
0.686& 
0.701 
\\ \hline

\multirow{3}{3em}{$c_3, t_3$} & 
bandit & 
\textbf{0.992}& 
0.950 & 
\textbf{0.986} & 
\textbf{0.976}& 
0.971 & 
\textbf{0.989} & 
\textbf{0.968} 
\\ \cline{2-9}

& 
sequential& 
0.967 & 
\textbf{1.000}& 
0.377& 
0.781 & 
\textbf{0.983}& 
0.672& 
0.689 
\\ \cline{2-9}

& 
random& 
0.959& 
0.995& 
0.412& 
0.789& 
0.977& 
0.686& 
0.704 
\\ \hline
\end{tabular}
\caption{Performance metrics for different algorithms in BEACON, across  configurations and timeframes. $c_1$ - 12/0/12, $c_2$ - 8/8/8, $c_3$ - 2/20/2, i.e., decreasing user constraints.}
\label{tab:results_table}
}
\end{table}
\section{Conclusion}
In conclusion, in this paper, we introduced   the novel problem of meal recommendation considering different meal configurations and time horizons,  presented the BEACON system which utilizes the boosted bandit method to address the problem of meal recommendation, displayed a dataset of 50 R3 items consisting of non-fast food and fast food items (Taco Bell and McDonald's), contributed a unique goodness metric that can be used to assess the quality of recommendations, and showed the efficacy of the boosted bandit method for generating robust recommendations across $3$ user configurations and $3$ time frames. We believe this can be a promising path towards promoting user adherence to dietary nutrition guidelines while balancing convenience.

In the future, one can extend this work in many directions, including: (1) implementing a more automated approach for generating R3 representations of recipes to increase the size of our dataset as this leads to more robust models; (2) increasing the number of features in terms of ingredients/allergens so that our dataset is more varied, and users with more allergens can receive positive recommendations; (3) experimenting with different recommendation algorithms and methods so that we may further explore the use of R3 representations;  (4) deploying an application that allows users to input their preferences and get a list of recommendations, and (5) conducting qualitative evaluation to show the acceptance of our recommendation system. 

\bibliographystyle{ACM-Reference-Format}
\bibliography{references}


\end{document}